\documentclass[conference]{IEEEtran}
\IEEEoverridecommandlockouts
\usepackage{cite}
\usepackage{amsmath,amssymb,amsfonts}
\usepackage{algorithmic}
\usepackage{graphicx}
\usepackage{textcomp}
\usepackage{xcolor}
\usepackage[pagebackref,breaklinks,colorlinks]{hyperref}
\def\BibTeX{{\rm B\kern-.05em{\sc i\kern-.025em b}\kern-.08em
    T\kern-.1667em\lower.7ex\hbox{E}\kern-.125emX}}
\begin{document}

\def\needtocite{\textcolor{red}{[CITATIONS]}}
\newcommand{\red}[1]{\textcolor{red}{#1}}

\title{
    A Novel Dataset for Video-Based Neurodivergent Classification Leveraging Extra-Stimulatory Behavior
}

\author{\IEEEauthorblockN{Manuel Serna-Aguilera}
\IEEEauthorblockA{\textit{Dept. of EECS} \\
\textit{University of Arkansas}\\
Fayetteville, United States \\
mserna@uark.edu}
\and
\IEEEauthorblockN{Xuan Bac Nguyen}
\IEEEauthorblockA{\textit{Dept. of EECS} \\
\textit{University of Arkansas}\\
Fayetteville, United States \\
xnguyen@uark.edu}
\and
\IEEEauthorblockN{Han-Seok Seo}
\IEEEauthorblockA{\textit{Dept. of Food Science} \\
\textit{University of Arkansas}\\
Fayetteville, United States \\
hanseok@uark.edu}
\and
\IEEEauthorblockN{Khoa Luu}
\IEEEauthorblockA{\textit{Dept. of EECS} \\
\textit{University of Arkansas}\\
Fayetteville, United States \\
khoaluu@uark.edu}
}

\maketitle

% Notes:
% IEEE Greentech is not double-blind review, so we can keep the author names visible in our submission

%****************************************
% Abstract
%****************************************
\begin{abstract}
Facial expressions and actions differ among different individuals at varying degrees of intensity given responses to external stimuli, particularly among those that are neurodivergent. Such behaviors affect people in terms of overall health, communication, and sensory processing. 
Deep learning can be responsibly leveraged to improve productivity in addressing this task, and help medical professionals to accurately understand such behaviors.
In this work, we introduce the Video ASD dataset—a dataset that contains video frame convolutional and attention map feature data—to foster further progress in the task of ASD classification.
Unlike many recent studies in ASD classification with MRI data, which require expensive specialized equipment, our method utilizes a powerful but relatively affordable GPU, a standard computer setup, and a video camera for inference. Results show that our model effectively generalizes and understands key differences in the distinct movements of the children.
Additionally, we test foundation models on this data to showcase how movement noise affects performance and the need for more data and more complex labels.
\end{abstract}

%\begin{IEEEkeywords}
%Deep Learning, Facial Expressions, Video, Classification
%\end{IEEEkeywords}

%****************************************
% One-page of greentech paper
%****************************************
\textbf{Introduction and Related Work. }
Deep learning has seen success in various deep learning tasks \cite{truong2022direcformer, nguyen2023brainformer, nguyen2023fairness, he2016deep, nguyen2023insect, nguyen2021clusformer, nguyen2023micron, nguyen2020self}, but recognizing atypical facial expressions is a relatively unexplored area. This can be attributed to the lack of a sufficiently-large dataset to help researchers better understand such behaviors. 
There are current works that use images, but not video \cite{image-asd-paper-2023} but do not make use of crucial temporal information for behavior analysis with Transformer models \cite{vit-paper-2020, swin-transformer-paper-2021, mehta2022mobilevit}. Further works analyze behaviors via video \cite{activity-recognition-asd-detection-2021}, gaze analysis \cite{Jiang_2017_ICCV, fang-gaze-cnn-lstm-2020, eye-track-asd-detetion-mazumdar-2021, eye-tracking-DL-asd-detection-ahmed-2022}, other behaviors \cite{Chen_2019_ICCV}, or via expensive equipment such as fMRI \cite{pyramid-kernel-asd-diagnosis-2023, hu-fmri-semi-supervised-lstm-asd-detection-2023, fmri-autoencoder-fscore-asd-detection-2023, fmri-asd-detection-cnn-2020, mri-asd-detection-rakic-2020, deep-svm-meta-learning-asd-diagnosis-2023, convolution-kernel-asd-detection-2021}, which may differ in degree between neurodivergent individuals.
By contrast, we analyze ASD-related behaviors by explicitly evoking reactions in more controlled settings.

\textbf{Data Collection. }
%Our dataset collection has been a collaboration between The University of Arkansas and The University of Texas at San Antonio. 
Each instance of our dataset contains a sequence of frame features for each video, thus obscuring the identify of the subjects, the ASD classification label (yes or no), and a text description. The frame features describe a person reacting to a sensory stimulus (e.g., taste or smell), with the video recording the subject's reaction in terms of the head pose (behavior), and facial expression. The text descriptions provide a description of the expression for every six second interval, e.g., ``The face expression changes from neutral to disgust'' if the face expression changes within that time frame.

\textbf{Methodology. }
Our method consists of two main components. First is the encoder, which is itself composed to two sub-encoder modules. We have one model (EfficientNet B0 CNN \cite{efficientnet-paper-2020}) to focus on neurotypical behaviors while the other model (ResNet-18 \cite{resnet-paper-2015} pretrained for the facial expression recognition task on MS-Celeb \cite{ms-celeb-dataset-paper-2016}) focuses on more neurodivergent behaviors. These deep learning-based models focus on spatial features. The input consists of 32 frames for each video. The second main component is the decoder, which is implemented by a spatial transformer, e.g., a vision transformer \cite{vit-paper-2020}. We then use a fully-connected network (or MLP) to output the probabilites for the classes. 
When we sample video slices, we average the probability predictions to obtain a final prediction.

\textbf{Experimental Results. }
% Implementation and hyperparams
We implement our framework using Pytorch \cite{pytorch-paper}.
We train and test our model using five $k$ folds, with four folds used as training samples, and the last fold for testing.
Our model trains with the AdamW optimizer \cite{adamw-optimizer-paper} with a cosine annealing scheduler \cite{cosine-scheduler-paper}, a learning rate 0.0001, and a cross entropy loss. We use a batch size of 4 on a Quadro RTX 8000 GPU with 48 GB of memory. As a preprocessing step, we perform face alignment via landmark detection and pose angle estimation is used to filter out bad frames. All input images have the spatial dimensions of $224 \times 224$.
For evaluation, we compute the classification accuracy at the per-video level.
The test accuracy on the test fold is 81.48\% with an F1 score of 0.7289. The result indicates the model is able to generalize well to similar but unseen samples, despite limited data and the number of frames processed per video. Despite this, such a low number frames may hinder the generalizability of the model, and movement noise of the subjects may contribute significantly to performance. Hence, we filter out frames where any head pose angle is out of the range $[-10, 10]$ to keep the data as controlled and noise-free as possible.
Future work will involve addressing how to incorporate frames with more extreme head poses, allowing for the analysis of up to hundreds of more frames with critical information, and account for other kinds of noise like movement and occlusions of the face.

%****************************************
% References
%****************************************
\bibliography{bib/egbib}{}
\bibliographystyle{plain}

\end{document}